\documentclass[report]{elsarticle}

\usepackage{array,xspace,multirow,hhline,tikz,colortbl,tabularx,booktabs,fixltx2e,amsmath,amssymb,amsfonts,amsthm}
\usepackage{paralist}
\usepackage[boxed]{algorithm}
\usepackage[noend]{algorithmic}

\usepackage[margin=2.5cm]{geometry}
\usepackage{amssymb,amsthm,xspace}

\newcommand{\ceil}[1]{\lceil #1 \rceil }

\usepackage{verbatim,ifthen}
\usepackage{enumitem}
\usepackage{url}
\usepackage{pifont}
\usepackage{ifthen}
\usepackage{calrsfs,mathrsfs}
\usepackage{bbding,pifont}
\usepackage{pgflibraryshapes}
\usetikzlibrary{arrows}
\usepackage{centernot}
\usetikzlibrary{decorations.pathreplacing}
\usetikzlibrary{patterns}
\usepackage{pgflibraryshapes}
\usetikzlibrary{positioning,chains,fit,shapes,calc}
\usepackage{varioref}
\usepackage{nicefrac}
\usepackage{hyperref}
\usepackage{cleveref}
\usepackage{tikz}
\usetikzlibrary{topaths,calc}
\usepackage[normalem]{ulem}
\usepackage{tabularx}
\usepackage{mathtools}

\definecolor{light-gray}{gray}{0.9}

\newtheorem{definition}{Definition}%

\usepackage{boxedminipage}
\usepackage{xspace}

\usepackage{subfigure}

\newcommand{\M}{{M}\xspace}
\newcommand{\N}{{N}\xspace}
\newcommand{\X}{\mathcal{X}\xspace}

\newcommand{\bt}[1][]{\ensuremath{\ifthenelse{\equal{#1}{}}{\mathit{BT}}{\mathit{BT}(#1)}}\xspace}

\newtheorem{lemma}{Lemma}%
\newtheorem{theorem}{Theorem}%
\newtheorem{proposition}{Proposition}%
\newtheorem{corollary}{Corollary}%
\newtheorem{example}{Example}

\newtheorem{remark}{Remark}

\usepackage{eqparbox}
\usepackage{enumitem}
\setenumerate[1]{label=\rm(\it{\roman{*}}\rm),ref=({\it\roman{*}}),leftmargin=*}
\newlength{\wordlength}

\newcommand{\midd}{\mathbin{:}}

\usepackage{enumitem}
\setenumerate[1]{label=\rm(\it{\roman{*}}\rm),ref=({\it\roman{*}}),leftmargin=*}

\newcommand{\nbh}[1][]{
	\ifthenelse{\equal{#1}{}}{\nu}{\nu(#1)}
}

\newcommand{\cstr}[1][]{
	\ifthenelse{\equal{#1}{}}{\mathscr S}{\cstr(#1)}
}

\newcommand{\choice}[1][]{
	\ifthenelse{\equal{#1}{}}{\mathit{C}}{\choice(#1)}
}
\newcommand{\jvnew}[1]{\textcolor{black}{#1}}

\newcommand{\nimrodout}[1]{}

\begin{document}
	
	%\date{}
	
	\title{Proportionally Representative Clustering}

\author{Haris Aziz}%\corref{cor1}}
\ead{haris.aziz@unsw.edu.au}
\address{UNSW Sydney}

\author{Barton E. Lee} 
\ead{bartonlee@ethz.ch}
\address{ETH Z{\"u}rich}

\author{Sean Morota Chu}%\corref{cor1}}
\ead{seanmorotachu@gmail.com}
\address{UNSW Sydney}

\author{Jeremy Vollen}%\corref{cor1}}
\ead{vollen@northwestern.edu}
\address{Northwestern University}

%
%\address{UNSW Business School,\\ Sydney, Australia}

%\renewcommand*{\today}{15 Oct 2022}

%	\today\\

\addtocounter{footnote}{0}

\begin{abstract}
	{In recent years, there has been a surge in effort to formalize notions of fairness in   machine learning. We focus on \emph{centroid} clustering}---one of the fundamental tasks in unsupervised machine  learning. 
	We propose a new axiom ``proportionally representative fairness'' (PRF) that is designed for clustering problems where the selection of centroids  reflects the distribution of data points and how tightly they are clustered together. 
	Our fairness concept is not satisfied by existing fair clustering algorithms. 
	We design efficient algorithms to achieve PRF both for unconstrained and discrete clustering problems. 
	Our algorithm for the unconstrained setting is also the first known polynomial-time approximation algorithm for the well-studied Proportional Fairness (PF) axiom \citep{CFLM19a} in general metric spaces. 
	Our algorithm for the discrete setting also matches the best known approximation factor for PF.
\end{abstract}

\maketitle

\section{Introduction}

As artificial intelligence and machine learning serve as the `new electricity' of systems, it is becoming increasingly  important to ensure that AI systems embody societal values such as privacy, transparency, and fairness.  Fairness is a    {growing} concern, as AI systems are used to make decisions that can critically affect our daily lives, finances, and careers~\citep{ChRo20a}. 
The impulse to tackle the issue of fairness in AI is prominently reflected in several governmental policy documents~\citep{Cong21a,Sele19a,Euro20a}, the emergence of dedicated conferences on Ethical AI, and the formation of AI ethics boards by various big tech companies. 
Over the past years, several developments have taken place in the theory and application of fairness in supervised learning. One  {prominent approach is}   to use     {information derived from} labelled data in supervised learning to formalize some   {notions} of equity within and across  {(pre-specified and)} protected attributes~\citep{DHP+12a}. In contrast, in several models of unsupervised learning, protected attributes or their labelling   {are} unknown. Fairness in unsupervised machine learning is becoming important with the growth of data collection both in public (e.g., via IoT devices) and private sectors (e.g., map services and social media). Unfettered and irresponsible use of such data has the danger  {of increasing} inequity~\citep{Neil16a}.

We focus on fairness in clustering, which is  one of the most widely applied  {tasks}  in unsupervised machine learning. 
A (centroid  {selection}) clustering problem has a metric space $\X$ with a distance measure $d:\X \times \X \rightarrow \mathbb{R}^+\cup \{0\}$, a multiset $\N\subseteq \X$ of $n$ data points (agents), a set $\M$ of candidate centers, and positive integer $k\leq n$.  {In many clustering problems, each data point corresponds to an individual's attribute profile; thus, we refer to the data points as \emph{agents}.}  The goal is  {to} find a  {size-$k$} subset  {of centers,} $Y\subseteq \M \ : \ |Y|=k$.  {Such \emph{centroid selection} problems can be widely applied to various applications including recommender systems and data-compression. The problem also captures social planning scenarios where $k$ facility locations need to be chosen to serve a set of agents.}

In such a (centroid  {selection}) clustering problem, there are two versions of the problem. In the \emph{discrete} version, { the set of candidate centers $\M$ is a finite subset of $\X$.} In the \emph{unconstrained} or \emph{continuous} setting, $\M=\X$. We   first focus on the unconstrained setting  as done by \citet{MiSh20a}.  {Later we discuss how our concepts and ideas extend to the discrete setting.} For a location $i$ and set of locations $S \subseteq \M$, we will denote $\min_{c\in S}d(i,c)$ as $d(i,S)$. 
{Standard clustering solutions include $k$-center, $k$-medians, and $k$-means.}\footnote{ {The $k$-center solution outputs a subset $Y\in \arg\min_{W\subseteq \M, |W|=k} \max_{i\in \N} d(i,W)$. The $k$-medians solution outputs a subset $Y\in \arg\min_{W\subseteq \M, |W|=k} \sum_{i\in \N} d(i,W)$. The $k$-means solution outputs a subset $Y\in \arg\min_{W\subseteq \M, |W|=k} \sum_{i\in \N} d(i,W)^2$.}}    {These  standard objectives can be viewed as achieving some form of global welfare but are not well-suited for proportional representational and fairness concerns.}  The problems of computing outcomes optimising these objectives are also NP-hard~\citep{MeSu84a,ADH+09a}.

We seek a suitable concept for clustering that captures general fairness principles such as non-discrimination, equality of opportunity, and equality of outcome. 
In many applications, the data points correspond to real individuals or their positions or opinions. 
Such individuals may expect the clustering  {outcome} to capture natural fairness requirements.
{
	Previous axiomatic studies of clustering have measured fair representation by the distance from each data point to the nearest centroid. 
	We initiate a proportional representation perspective on fairness in clustering.
	In capturing this desiderata axiomatically, we require a property in which representation guarantees depend on the number {(or proportion)} of data points and how ``tightly'' they are clustered together.
	
	Our perspective is   motivated by applications in which groups of individuals (to which the data points correspond) may expect to receive centroid representation proportional to their size.
	For instance, the centroids may be citizens selected to make representative decisions on behalf of the public, as is done in a jury or a sortition panel.
	Similarly, computing a proportionally fair clustering is also meaningful in facility location, where the location of the facility should depend on the density and the number of people it is serving. 
	More generally, our axioms and algorithms are well-motivated in any setting in which one may want to take a representative sample from datapoints between which a distance metric is well-defined.
}
In this paper, we consider the following fundamental research question. \emph{	What is a meaningful concept of proportional fairness in centroid clustering? Is such a concept  guaranteed to be achievable? }

\paragraph{Contributions} In contrast to traditional optimization objectives in clustering and centroid selection, we take an axiomatic fairness perspective. 
We first identify a potential shortcoming of some recently introduced concepts that  {aim to capture proportional fairness for clustering;}  in particular, we show how the concepts do not capture{ an intuitive and minimal requirement of proportional representation called \emph{unanimous proportionality}.} We then propose a new  {axiom} called \emph{proportional representative fairness (PRF)}, which is our central conceptual contribution.  

We propose PRF  {for both the unconstrained and discrete clustering setting.}   
We discuss how PRF overcomes some of the drawbacks of the previously introduced concepts. 
In particular, it implies{ the minimal requirement of} unanimous proportionality.
PRF has several {other} desirable features{: a PRF outcome is guaranteed to exist, the axiom does not require a}    preset specification of protected groups  and  {it is} robust to outliers and multiplicative scaling. We show that none of the traditional clustering algorithms {or} previously introduced algorithms for fair clustering satisfy PRF. 

We design polynomial-time algorithms to achieve PRF both in the unconstrained  {and} discrete settings.\footnote{The results for unconstrained and discrete settings are incomparable in the following sense.  {The guarantee of  existence of a fair outcome in the discrete setting  guarantees the existence of a fair outcome in the unconstrained setting; however, a polynomial-time algorithm for computing a fair outcome in the discrete setting does not imply the same for the unconstrained setting.} } 
{We call our algorithms \textit{SEAR (Spatial Expanding Approval Rule)} as they are inspired by using core ideas from the ordinal multi-winner voting Expanding Approvals Rule (EAR) of \citet{AzLe19a} but carefully use  spatial information to make synchronized decisions.}
{We prove that our algorithm for the unconstrained setting also gives $3$-approximate \textit{Proportional Fairness (PF)}~\citep{CFLM19a}, and hence is the first known polynomial-time approximation algorithm for PF on general metric spaces.}
{We also prove that our algorithm for the discrete setting matches the best known approximation of \textit{Proportional Fairness (PF)}, thereby providing no compromise on PF approximation but additionally satisfying PRF.} 
We summarize our theoretical contributions in Table~\ref{tab:results_summary} via a comparison of our algorithms with the best guarantees previously known.

Finally, we show that there is a tradeoff between strategyproofness and PRF properties. 
%All missing proofs can be found in \ref{appendix:proofs}. 
% {Finally, w}e experimentally compare our algorithm with two standard algorithms and show that it performs especially well when the mean squared distance of the agents is  {calculated based on their distance from not only the closest center but also the next $j$-closest centers (for large $j$).}

%establish further logical relations between approximations of various fairness concepts. In particular, 

%
\begin{table}[t]
	\centering
	\scalebox{1}{
		\addtolength{\tabcolsep}{0.2em}
		\begin{tabular} {@{}rrcrcrcr@{}} \toprule	& \multicolumn{3}{c}{Unconstrained} & \phantom{abc}& \multicolumn{3}{c}{Discrete}\\
			\cmidrule{2-4} \cmidrule{6-8}
			& $\rho$-PF & PRF & Complex. && $\rho$-PF & PRF & Complex. \\ \midrule
			% Our algorithms
			\textbf{Our Algorithms} ({SEAR}) & $\mathbf{3}$ & \checkmark & \textbf{in P} && $\mathbf{\approx2.4}$ & \checkmark & \textbf{in P} \\
			% Existing
			Greedy Capture & $\approx2.4$ & - & NP-hard && $\approx2.4$ & - & in P \\
			% (best known) & &  & && & & \\
			\bottomrule
		\end{tabular}
	}
	\caption{
		Theoretical comparison of our algorithms (Algorithm~\ref{alg:PRF-unconstrained} for the unconstrained setting and Algorithm~\ref{alg:PRF-discrete} for the discrete setting) with Greedy Capture (GC), which gives the best known approximate PF guarantees. Unconstrained and discrete GC results come from \protect\citet{MiSh20a} and \protect\citet{CFLM19a}, respectively.}
	%Note that \citet{MiSh20a} provide a ($2+\epsilon$)-PF PTAS, but this result does not hold for general metric spaces.}
\label{tab:results_summary}
\end{table}

% \footnotetext{Unconstrained and discrete Greedy Capture results come from \citet{MiSh20a} and \citet{CFLM19a}, respectively.}

\section{Related Work}
\subsection{Fair clustering}
Fairness in machine learning and computer systems  is a vast topic~\citep{FrNi96a}{---for an overview, see  \citet{ChRo20a} or the survey by \citet{MMS+21a}.} We focus on the fundamental learning problem of clustering, which is a well-studied problem in computer science, statistics, and operations research~\citep{ELL09a}.  {Our approach is axiomatic, focused on the concept of proportional fairness (or representation), and does not rely on the use of a pre-specified or protected set of attributes. Thus, the most closely related papers are those of  \citet{CFLM19a}, \citet{MiSh20a}, \citet{LLS+21a}, and \citet{JKL20a}, which we review below.}

\citet{CFLM19a} proposed the concept of \emph{proportional fairness}  {(PF)} for the discrete setting, which is based on the entitlement of groups of agents of large enough size. They reasoned that  {PF}  is a desirable fairness concept but it is not guaranteed to be achievable for every instance. On the other hand, they showed that outcomes satisfying reasonable approximations of    {PF} are guaranteed to exist and can be computed in polynomial time via the `Greedy Capture' algorithm.

\citet{MiSh20a}   {analyzed the} proportional fairness  {axiom} in the  unconstrained setting and presented  similar results  {to those obtained by \citet{CFLM19a}}. 
They showed that the natural adaptation of the Greedy Capture algorithm finds a 2-approximation to PF in Euclidean space, but cannot be computed in polynomial time unless P=NP.
However, they give a $(2+\epsilon)$-PF polynomial time approximation algorithm for Euclidean space.
As this algorithm uses a PTAS from computational geometry which applies only to Euclidean spaces, it does not easily extend to general metric spaces.
In this work, we give the first known polynomial time approximation algorithm for proportional fairness in general metric spaces.

\citet{LLS+21a} consider the same discrete clustering setting as \citet{CFLM19a} and propose a stronger version of proportional fairness called \emph{core fairness.} {They}  present   results on lower and upper approximation bounds for various restrictions of the discrete setting.   
%{Because} core fairness is stronger than proportional fairness, the negative existence results of  \citet{CFLM19a} also hold for core fairness.

%In a paper inspired by our work, \citet{KePe23a} analyze the key algorithms and axioms in our paper in further detail and show that the axioms that we propose also imply approximations of various fairness concepts introduced in the literature. 
%Almost all of their  key results for constructively satisfying combinations of fairness guarantees are achieved via the algorithms that we present and formalize for the first time in the spatial context. 
%A key takeaway from our paper is reaffirmed by their work, namely that our algorithms are the best-known algorithms for achieving proportional representation axioms and corresponding approximations.

\citet{JKL20a} consider the discrete clustering problem in which the set of agents and candidate centers coincide. For  {this} setting, they propose a fairness concept that  has a similar goal as that of \citet{CFLM19a}. In subsequent work, it has been referred to as \emph{individual fairness}~\citep{VaYa21a,ChNe21a}.  
{Individual fairness} requires that, for each agent $i$, there  {is} a selected center that is at most $r(i)$ from $i$, where $r(i)$ is the smallest radius around $i$ that includes $\ceil{n/k}$ agents.   

In a paper inspired by our work, \citet{KePe24b} analyze the key algorithms and axioms in our paper in further detail and show that the axioms that we propose also imply approximations of various fairness concepts introduced in the literature. 
\jvnew{They also formalize a connection between approval-based multiwinner voting and the centroid-based clustering setting, letting each instance of centroid-based clustering and threshold value $y$ correspond to a unique instance of approval-based multiwinner voting wherein voters approve of exactly those candidates within distance $y$.
Then, the metric analog of an approval-based axiom simply requires that the latter be satisfied in the appproval-based instance resulting from any threshold value $y$.
According to this translation, they show that the axiom we introduce in \Cref{def:prf_discrete} is analogous to the PJR axiom from multiwinner voting.}
% Drawing an analogy to multiwinner voting axioms, they refer to our proposed axiom PRF for the discrete setting as mPJR and also define a weaker axiom which they call mJR.
They then do a deep dive into the logical relations between the axioms and approximate notions of proportional fairness, individual fairness, core fairness, and another core-inspired generalizations of proportional fairness \citep{EbMi24a}. 
% They prove that any approximation to PF is also an approximation to individual fairness and vice versa.
In addition to the algorithms we formalize in this paper, they also consider Greedy Capture \citep{CFLM19a} and a randomized version of this algorithm known as Fair Greedy Capture \citep{EbMi24a}.
A key takeaway from our paper is reaffirmed by their work, namely that our algorithms are the best-known algorithms for satisfying proportional representation axioms while also giving non-trivial approximation guarantees with respect to other fairness notions.

\jvnew{Like each of the papers mentioned above, our work focuses on centroid clustering, but subsequent work has examined non-centroid clustering from a fairness perspective \citep{CMS24a,CSY25a,BDK+25a}.
In particular, \citet{CMS24a} study a setting in which an agent's preferences are a function of the other agents in their cluster, rather than the centroids selected. 
They are able to adapt the Greedy Capture algorithm of \citet{CFLM19a} to give positive results for core-related fairness notions in their setting.
\citet{CSY25a} later studied a setting between that of centroid and non-centroid clustering, which they call semi-centroid clustering.}

There are also several concepts and algorithms in clustering where the protected groups are   {pre-specified} based on protected attributes~\citep{ABV21a,AEK+19a,BCFN19a,BIO+19a,KSAM19a,CKLV17a,CKR20a,EBS21}. In contrast, we are seeking fairness properties that apply to arbitrarily defined groups of all possible sizes.  {In our proposed axiom, these groups are \emph{endogenously} determined from data points; in some real-world  settings, this is a legal requirement since it can be forbidden by law to have decision-making processes use certain attributes as inputs}.  Clustering can also be  viewed as a facility location problem~\citep{FaHe09a}, where multiple facilities (centers) are to be located in a metric space; for a recent overview, see the survey by \citet{CFL+21}.

\subsection{Comparison with multi-winner voting}
The clustering problem especially has natural connections with multi-winner voting in which agents and centers in the clustering context  correspond  to agents and candidates in multi-winner voting.

{
The literature on proportionally representative multi-winner voting includes axioms such as JR, PJR, EJR, and FJR for dichotomous preferences~\citep{ABC+16a,LaSk23a}, axioms such as PSC for rankings~\citep{Dumm84a}, and axioms such as Generalised PSC~\citep{ABC+16a,AzLe19a} or Rank-PJR+\citep{BrPe23a} for weak orders.
Naively applying these axioms to our spatial problem results in a reduction from our unconstrained problem to a multiwinner voting problem with an infinite number of candidates. 
Conversely, our fairness (or proportional representation) axioms take into account the additional distance-based {cardinal} information available when the candidates belong to a general metric space. 
Hence, our axioms formalize a notion of  proportional representation for $m$-dimensional Euclidean space for $m>2$. 
In contrast to ordinal algorithms for multi-winner voting ({such as the Expanding Approvals Rule (EAR)~\citep{AzLe19a} or Single Transferable Vote}), the algorithms in this paper carefully account for additional spatial information to make synchronized decisions.

{\citet{KKK24a} consider a similar setting to ours but explore to what extent algorithms that are agnostic to cardinal spatial information and only use ordinal ranking of pairwise distances can achieve approximate fairness.}
{They define $(\alpha,\gamma)$-proportional representation, a powerful strengthening of approximate core fairness~\citep{LLS+21a}.
	Their notion captures (and strengthens) the ideal of core stability in this setting: no coalition of agents $S$ large enough to ``deserve'' $t$ centroids should be able to propose an alternative set of $t$ centroids 
	which they all prefer to their individually best size-$t$ subset of the selected centroids.
	They then parameterize this idea by $\alpha$, the required size of a coalition which deserves $t$ centroids, and $\gamma$, the degree to which the alternative centroid selection better represents the coalition.
	Similar to ours, their property effectively addresses the shortfalls of existing concepts with respect to proportional representation (see Definition~\ref{def:up}).
	In contrast to theirs, the properties we study do not require parameterization and are guaranteed to exist.}
{
	\citet{KKK24a} show that the (ordinal multi-winner) Expanding Approvals Rule (EAR) of \citet{AzLe19a} is $(\alpha,\gamma)$-representative with $\gamma \leq  1 + 6.71 \alpha/(\alpha-1)$ and also satisfies $5.71$-approximation to proportional fairness.}
{They are able to improve on these bounds with a modified version of Greedy Capture when cardinal information is known.}

Proportionally representation in multi-winner voting (and more generally participatory budgeting) has also been studied when agents have cardinal utilities. 
{Based on prior work for approvals~\citep{ABC+16a}, \citet{PPS21a} study desirable axioms such as FJR and EJR for positive, \textit{additive} utilities, and rules such as the \textit{Method of Equal Shares} (MES).
	We make no assumption of additive utilities, nor do we ever consider total or average distances. 
	Under additive cardinal utilities, a group of agents of proportion large enough to “deserve” multiple candidates can (in some instances) be satisfied with respect to EJR by a single candidate which gives some agent very high utility. 
	Because our model and axioms take an agnostic approach to cardinal utility values, we instead place requirements on the selection of a proportional number of candidates. 
	One possible approach to our problem is to take distances as proxy for utilities and to apply a rule like MES.
	However, translating to cardinal utilities necessitates the loss of important spatial information as different cardinal utilities consistent with the distances give different MES outcomes. 
	One possibility is to set utility to be negation of the distance. However, the setting and axioms of Peters et al. assume {\textit{positive}}, additive utilities.} 
	Another possibility is to set utility to be the inverse of the distance. \jvnew{Under this transformation, MES still fails our introduced axiom, which we show in~\ref{app:mes_violates_prf}.}

%\section{Limitation of Existing Proportional Fairness Concepts}
\section{Towards an Axiom Capturing Proportional Representation}

Proportional fairness for clustering has been considered in a series of  {recent} papers. \citet{CFLM19a} first proposed  {a} proportional fairness  {axiom} that requires that there is no set of agents of size at least \jvnew{$\ceil{n/k}$} that can  {find}   a{n unselected} center  that is ``better''  {(i.e., located  closer than the closest selected center)} for each of them. 

\begin{definition}[Proportional Fairness~\citep{CFLM19a}]
	$X \subseteq \M$ with $|X| = k$ satisfies proportional fairness if $\forall S \subseteq \N$ with $|S| \geq \lceil \frac{n}{k} \rceil$ and for all $c \in \M$, there exists $i \in S$ with $d(i, c) \geq d(i,X)$. 
\end{definition}

The idea of proportional fairness was further strengthened to core fairness that requires that there is no set of agents of size at least \jvnew{$\ceil{n/k}$} that can demand a more preferred location for a center~\citep{LLS+21a}.
One rationale for proportional fairness is explained in the following example{, which was} first presented by \citet{MiSh20a} and then reused by \citet{LLS+21a}. We reuse the same example. 
Consider a set of data points/agents. The agents are divided into 11 subsets of clusters each of which is densely clustered. One cluster of agents has size 10,000. The other 10 clusters have sizes 100 each. The big cluster is very far from the smaller clusters. The small clusters are relatively near to each other.
\citet{MiSh20a} and \citet{LLS+21a} are of the view that any set of centers that satisfies a reasonable notion of proportional fairness would place 10 centers for the big cluster and 1 center serving the small clusters. Next, we point out that proportional fairness and core fairness do not necessarily satisfy this minimal requirement{, which we formalize via the \emph{unanimous proportionality} axiom below.}  

% \medskip
\begin{definition}[Unanimous Proportionality (UP)]	\label{def:up}
	$X \subseteq \M$ with $|X| = k$ satisfies \emph{unanimous proportionality} if  $\forall S \subseteq \N$ with $|S| \geq \ell \ceil{\frac{n}{k}}$ and each agent in $S$ has the same location $x\in \X$, then $\ell$ possible locations closest to $x$ are selected.
\end{definition}

UP captures a principle which \citet{CFLM19a} highlighted as desirable --- in their words, one which requires that \emph{``any subset of at least $r\ceil{n/k}$ individuals is entitled to choose $r$ centers.''}\footnote{In the term unanimous proportionality, unanimous refers to the condition where a set of agents have the same location, and proportionality refers to the requirements that such agents get an appropriate number of centers in proportion to the number of agents.}
The following example shows that proportional fairness~\citep{CFLM19a} does not imply unanimous proportionality.
%, and nor does core fairness concept of \citet{LLS+21a}. 
Similarly, it can be shown that the fairness concept proposed by \citet{JKL20a} does not imply unanimous proportionality. 

\begin{example}[Proportional fairness and core fairness]
	Suppose agents are  {located} in the $[0,1]$ interval and $k=11$. 10,000 agents are at point 0 and 1000 agents are at point 1. 
	% Suppose we want to find $k=11$ centers.    
	{UP} requires that 10 centers are at or just around point 0 and 1 center is at point 1. However, placing 1 center at point 0 and 10 centers at point 1 satisfies the proportional fairness concept of \citet{CFLM19a} and also the core fairness concept of \citet{LLS+21a}.
\end{example}

The example above shows that there is a need to formalize new concepts in order to capture proportional representative fairness. 
The property should imply unanimous proportionality but should also be meaningful if no two agents/data points completely coincide in their location. 
% {This can be important for certain tasks such as data abstraction and sortition, as discussed in the introduction.}
{One reason why the existing fairness concepts do not capture UP is that they make an implicit assumption that %may be suitable in some contexts of facility location but possibly not in the context of certain unsupervised learning problems. The assumption is that 
	an agent \emph{only} cares about the distance from the nearest center and not on how many centers are nearby.} 
As discussed in the introduction, this assumption is not necessarily accurate in contexts such as data abstraction, sortition, and political representation.

In addition to capturing proportional representation in a robust sense, we would also like our axiom (and algorithms) to guarantee the existing proportional fairness concepts, at least in an approximate fashion. 
Since a PF outcome may not exist, \citet{CFLM19a} and \citet{MiSh20a} studied an approximate notion of proportional fairness in the discrete and unconstrained settings, respectively. 
% We will use this approximate notion in benchmarking our axiom and algorithms with respect to proportional fairness. 

\begin{definition}[$\rho$-approximate Proportional Fairness]
	$X \subseteq \M$ with $|X| = k$ satisfies $\rho$-approximate PF (or $\rho$-PF) if $\forall S\subseteq N$ with $|S|\geq \lceil \frac{n}{k} \rceil$ and for all $c \in \M$, there exists $i \in S$ with $\rho \cdot d(i,c) \geq d(i,X)$.
\end{definition}

We seek to devise an axiom which is compatible with approximate PF, and subsequently to design algorithms which satisfy our axiom while also providing good approximate guarantees with respect to PF.
We close the section with an example presented by  \citet{MiSh20a} that shows that a PF outcome may not exist. We will use the same example to illustrate our new fairness concept in the next section. 
\begin{example}\label{example:PFnott}
	Consider Figure~\ref{fig:PFnott}. For the $n=6$ agents and $k=3$, it has been shown by \citet{MiSh20a} that a PF outcome may not exist. \jvnew{We summarize the argument for the sake of completeness. Since each pair of agents can demand a dedicated center, PF requires that each agent among $\{1,2,3\}$ expects a center at distance at most $d(1,2)/2$ from it. This requirement can only be satisfied by 2 centers for agents $\{1,2,3\}$. By symmetry, we also need 2 centers for agents $\{4,5,6\}$. But this contradicts that we can only have $k=3$ centers.}
\end{example}

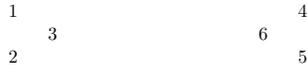
\begin{figure}[h!]
	\centering
	\scalebox{1.0}{
		\begin{tikzpicture}
			\node[] at (0, 0) (0) {};

			\node (1) at ([shift=({120:0.5 cm})]0) {1};
			\node (2) at ([shift=({240:0.5 cm})]0) {2};
			\node (3) at ([shift=({360:0.5 cm})]0) {3};

			\node[] at (5, 0) (0b) {};

			\node (5) at ([shift=({300:0.5 cm})]0b) {5};
			\node (4) at ([shift=({60:0.5 cm})]0b) {4};
			\node (6) at ([shift=({180:0.5 cm})]0b) {6};

		\end{tikzpicture}
	}
	\caption{An example instance with 6 agents and $k=3$ for which a PF outcome does not exist. }
	\label{fig:PFnott}
\end{figure}

\section{Fairness for the Unconstrained Setting}
\label{sec:fairness_unconstrained}

We propose a new concept \emph{Proportionally Representative Fairness (PRF)} that captures proportional representation fairness concerns.  The intuition behind the concept is based on two natural principles:
(1) a set of agents that is large enough deserves to have a proportional number of centers `near' it, and (2) the requirement of proximity of the nearby centers to a subset of agents should also  depend on how densely together the particular subset of agents are.
We use these two principles to propose a new fairness concept.  
\begin{definition}%[Proportionally Representative Fairness (PRF)\\ for Unconstrained Clustering]
	\emph{	\textbf{(Proportionally Representative Fairness (PRF) for Unconstrained Clustering)}}
	An outcome $X \subseteq \M$ with $|X| = k$ satisfies \emph{Proportionally Representative Fairness (PRF)} if the following holds. 
	For any set of agents $S\subseteq \N$ of size at least $|S|\geq \ell n/k$ such that the maximum distance between pairs of agents in $S$ is $y$, the following holds: 
	$|c\in X\midd \exists i\in S \text{ s.t } d(i,c)\leq y|\geq \ell$.
\end{definition}

PRF has the following features:
it implies unanimous proportionality; it is oblivious to and does not depend on the specification of protected groups or sensitive attributes; it is robust to outliers in the data, since the fairness requirements are for groups of points that are sizable enough; and it is scale invariant to multiplication of distances (i.e., PRF outcomes are preserved if all the distances are multiplied by a constant).

Next, let us illustrate an example where proportional fairness is not achievable but one of the most natural   {clustering} outcomes satisfies PRF.

\begin{example}[Requirements of PRF]
	We revisit Example~\ref{example:PFnott}    {to illustrate the}  requirements   {that} PRF imposes.  
	For each set $S$ of size at least \jvnew{$\ell n/k = 2\ell$},    {PRF requires} $\ell$ centers in the relevant neighborhood of agents in $S$ (see Figure~\ref{fig:PRF-}).  {Therefore, f}or agent 1 and 2, PRF requires that one selected candidate  {center} should be in one of the circles around 1 or 2. For the set of agents $\{1,2,3\}$, PRF also requires that one candidate  {center} should be in one of the circles around 1, 2 or 3.  {A natural solution is to have   one  center  located between agents 1, 2, and 3, another  center located between agents 4, 5, and 6, and the final center located somewhere not too far from all agents (e.g., between  the two groups of agents). Such solutions are intuitively reasonable and fair and, furthermore, satisfy PRF; however, as mentioned in Example~\ref{example:PFnott}, this instance does not admit a proportionally fair solution.}
	%PRF also requires that all three selected candidates should not be overl 
\end{example}

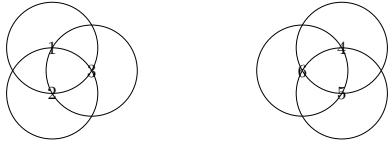
\begin{figure}[h!]
	\centering
	\scalebox{0.75}{
		\begin{tikzpicture}
			\node[] at (0, 0) (0) {};
			% \node at (120:1cm) (1) {1};
			%   \node at (240:1cm) (2) {2};
			%    \node at (360:1cm) (3) {3};
			
			\node (1) at ([shift=({120:0.5 cm})]0) {1};
			\node (2) at ([shift=({240:0.5 cm})]0) {2};
			\node (3) at ([shift=({360:0.5 cm})]0) {3};
			
			\draw[] (1) circle [radius=0.86602540378];
			\draw[] (2) circle [radius=0.86602540378];
			\draw[] (3) circle [radius=0.86602540378];

			\node[] at (5, 0) (0b) {};
			% \node at (120:1cm) (1) {1};
			%   \node at (240:1cm) (2) {2};
			%    \node at (360:1cm) (3) {3};
			
			\node (5) at ([shift=({300:0.5 cm})]0b) {5};
			\node (4) at ([shift=({60:0.5 cm})]0b) {4};
			\node (6) at ([shift=({180:0.5 cm})]0b) {6};
			
			\draw[] (4) circle [radius=0.86602540378];
			\draw[] (5) circle [radius=0.86602540378];
			\draw[] (6) circle [radius=0.86602540378];
			
		\end{tikzpicture}
	}
	\caption{Some of the requirements of PRF for instance in Example~\ref{example:PFnott}.}
	\label{fig:PRF-}
\end{figure}

{In addition to capturing natural outcomes when PF outcomes may not exist, PRF actually guarantees a $\rho$-PF outcome for $\rho\approx 3.6$ in the unconstrained setting on general metric spaces.}

\begin{proposition} \label{prop:approx_pf_unconstrained}
	{{For metric spaces,} if an outcome $X\subseteq \M$ with $|X|=k$ satisfies PRF for Unconstrained Clustering, then $X$ is $\frac{3+\sqrt{17}}{2}$-approximate PF, and there exists an instance for which this bound is tight.}
\end{proposition}
\begin{proof}
	Let $X$ be an outcome which satisfies PRF for Unconstrained Clustering.
	Fix $S\subseteq N, c\in \M$ with $|S| \geq \ceil{n/k}$.
	%First note that if $d(i,c) = 0$ for any $i\i0n S$, then $X$ satisfies PF.
	%Thus, we assume $d(i,c)>0$ for all $i\in S$.
	
	Let $y$ be the maximum distance between pairs of agents in $S$, i.e., $y=\max_{i,j\in S} d(i,j)$, and let $i_1$ and $i_2$ be two agents in $S$ that maximize this distance. 
	We assume that $y > 0$; otherwise, if $y=0$, then every agent in $S$ is located at the same point, and there exists $x\in X$ such that $d(i,x)=0\leq d(i,c)$ for all $i\in S$.
	Denote by $i^*$ the agent in $S$ for which the distance to $c$ is maximized. We have that 
	$d(i^*, c) \geq \max(d(i_1,c), d(i_2,c)) \geq \frac{1}{2}(d(i_1,c) + d(i_2,c)) \geq \frac{y}{2}$.
	
	Next, since \jvnew{$|S|\ge \ceil{n/k} \ge n/k$} and every pair of agents in $S$ are a distance of at most $y$ from each other, it follows from $X$ satisfying Unconstrained PRF that there exist $x\in X, i\in S$ such that $d(i,x)\leq y$.
	Using this along with the triangle inequality, we can establish the following upper bound: $d(i^*, x) \leq d(i^*, c) + d(c,x) \leq d(i^*,c) + d(i,c) + y$. 
	From these observations, the minimum PF approximation ratio obtained by $i$ or $i^*$ becomes
	\begin{align*}
		& \min \left ( \frac{d(i,x)}{d(i,c)}, \frac{d(i^*,x)}{d(i^*,c)} \right) \\
		& \leq \min \left ( \frac{y}{d(i,c)}, \frac{d(i^*, c) + d(i,c) + y}{d(i^*,c)} \right) \\
		& \leq \min \left ( \frac{y}{d(i,c)}, 1 + \frac{d(i,c)}{y/2} + \frac{y}{y/2} \right) \\
		& \leq \min \left ( \frac{y}{d(i,c)}, 3 + \frac{2\cdot d(i,c)}{y} \right) \\
		& \leq \max_{z\geq 0}(\min(z, 3 + 2/z)) = \frac{3 + \sqrt{17}}{2}.
	\end{align*}
	
	Note that the final inequality above holds since replacing $\frac{y}{d(i,c)}$ by the non-negative number $z$, which maximizes the expression, can only weakly increase the expression. It can be easily verified that this expression is maximized when $z=\frac{3+\sqrt{17}}{2}$, hence the final transition. Thus, for any $c\in \M$, there exists $i\in S$ such that $d(i,X) \leq \frac{3 + \sqrt{17}}{2} d(i,c)$. 
	%We defer the proof that this bound is tight to \ref{appendix:tightness}.
\end{proof}

The following example shows the analysis in Proposition~\ref{prop:approx_pf_unconstrained} is tight. That is, there is an instance for which an outcome satisfying PRF for Unconstrained Clustering obtains no better than $\frac{3+\sqrt{17}}{2}$-PF.

\begin{example}
	Consider an instance of unconstrained clustering with $k=1$ and $\N=\{i_1,i_2,i_3\}$. Furthermore, let $c$ and $x$ be two candidate locations in $\M$ such that the distance between agents and candidates is as given in Table~\ref{tab:tight_pf_unconstrained}.
	
	\begin{table}
		\centering
		\begin{tabular}{||c| c c c |}
			\hline
			& $i_1$ & $i_2$ & $i_3$ \\  
			\hline
			$i_1$ & 0 & $6+2\sqrt{17}$ & $7+\sqrt{17}$  \\ 
			\hline
			$i_2$ & $6+2\sqrt{17}$ & $0$ & $7+\sqrt{17}$ \\
			\hline
			$i_3$ & $7+\sqrt{17}$ & $7+\sqrt{17}$ & $0$ \\
			\hline
			$c$ & $3+\sqrt{17}$ & $3+\sqrt{17}$ & $4$ \\ 
			\hline
			$x$ & $13+3\sqrt{17}$ & $13+3\sqrt{17}$ & $6+2\sqrt{17}$ \\
			\hline
		\end{tabular}
		\caption{Distances between agents $\{i_1,i_2,i_3\}$ and candidates $\{c,x\}$.}
		\label{tab:tight_pf_unconstrained}
	\end{table}
\end{example}

It can be verified that all distances in Table~\ref{tab:tight_pf_unconstrained} satisfy the triangle inequality. 
Note that, since $6+2\sqrt{17}$ is the maximum distance between any pair of agents in $\N$ (call this $y$), and $d(i_3,x)=y$, it holds that $\{x\}$ satisfies PRF for Unconstrained Clustering.
Through straightforward arithmetic, it can then be shown that $\frac{d(i_1,x)}{d(i_1,c)}=\frac{d(i_2,x)}{d(i_2,c)}=\frac{d(i_3,x)}{d(i_3,c)}= \frac{3+\sqrt{17}}{2}$.

Next, we highlight that the {well-known} $k$-means, $k$-median, and $k$-center solutions do not necessarily satisfy PRF.   
{Example~\ref{example: k means no PRF} is adapted from an example by \citet{CFLM19a}.}

\begin{example}[\jvnew{$k$-means, $k$-median, and $k$-center do not satisfy PRF}]\label{example: k means no PRF}
	Consider Figure~\ref{fig:kmeans} in which there are $n/3$ agents uniformly distributed on the perimeter of each of the three circles.  
	\jvnew{If $k=3$, any $k$-means, $k$-median, or $k$-center solution will place one center between the small circles and two centers in the big circle, provided the big circle is sufficiently large.}
	In contrast, PRF requires that each of the circles gets its own centroid. 
	% It can also be shown that $k$-center does not satisfy PRF.

	\begin{figure}[h!]
		\centering
		\scalebox{0.75}{
			\begin{tikzpicture}

				\node () at (-0.5, +1.3) () { $n/3$ points};
				\node (1) at (0, +0.5) (1) {};
				
				\node () at (-0.5, -1.4) () { $n/3$ points};
				\node (2) at (0, -0.5) (2) {};

				\node () at (5.2, 1.4) () { $n/3$ points};
				\node (3) at (6, 0) (3) {};
				
				\draw[] (1) circle [radius=0.3];
				\draw[] (2) circle [radius=0.3];
				\draw[] (3) circle [radius=1];

			\end{tikzpicture}
		}
		\caption{The k-means solution may not satisfy PRF. }
		\label{fig:kmeans}
	\end{figure}
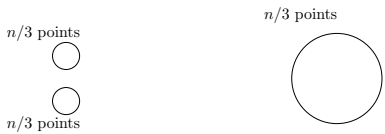
	
\end{example}

On face value, it is not clear whether an outcome satisfying PRF exists as similar concepts such as PF cannot always be guaranteed~\citep{MiSh20a}.  
Secondly, even if a PRF outcome is guaranteed to exist, the complex constraints seem computationally challenging. 
PRF is a property that requires conditions on an exponential number of subsets of agents. 
For each subset of agents, it enforces representational constraints pertaining to an infinite number of neighborhood distances.
{Perhaps surprisingly,}  we show that not only is a PRF outcome guaranteed to exist, it can be computed in polynomial time.

The algorithm intuitively works as {follows}. Firstly, instead of considering infinite possible centers, we restrict our attention to the $n$ candidate {centers  that coincide with each agent's location}. Each agent is given an initial weight of 1. This weight  {dynamically}  decreases over the course of the algorithm. 
The algorithm can be viewed as gradually expanding the neighborhoods of each  {of} the agents.\footnote{The idea of expanding neighborhoods is used in other algorithms in the literature, such as the Greedy Capture algorithm for the unconstrained setting studied by ~\citet{MiSh20a}. Unlike our algorithm, it does not involve reweighting of agents. More importantly, Greedy Capture does not satisfy PRF \jvnew{(see \Cref{rmk:prf_gc}). It is also not {polynomial-time} computable for the unconstrained setting.} {Our algorithm uses reweighting of agents as is done by EAR for multi-winner voting \citep{AzLe19a}}. Our algorithm can be viewed as using key ideas from the EAR for committee voting while taking into account additional spatial information to make synchronized decisions.}
Instead of continuously expanding the neighborhood, we only consider at {most} $n^2$ possible values of neighborhood radii. If there exists some location in the candidate set such that it is in the  {intersection of the} neighborhoods of  {a group of} agents with total weight  {of}  at least $n/k$, we select one such candidate, and reduce the agents' collective weight by $n/k$.\footnote{{We point out that the reweighting process can be done in any manner that results in the group's total weight reducing by $n/k$. Two natural possibilities would be (i) reducing weights as equally as possible, or (ii) reducing weights proportional to agent distances from the selected candidate.} \jvnew{The axiomatic results that we prove in the paper regarding SEAR do not hinge on which reweighting process is used.}} The process is continued until $k$ centers are chosen. The algorithm is formally specified as Algorithm~\ref{alg:PRF-unconstrained}. The following is the main result of the present section and summarizes the theoretical guarantees of Algorithm~\ref{alg:PRF-unconstrained}. 

\begin{theorem} \label{thm:unconstrained}
	Algorithm~\ref{alg:PRF-unconstrained} terminates in polynomial time \jvnew{$O(n^2\log n+kn^2)$} and returns a set of $k$ centers which satisfy PRF and 3-approximate PF. 
\end{theorem}

We will now formally prove two key lemmas used to prove Theorem~\ref{thm:unconstrained}.
% We defer the proof of Theorem~\ref{thm:unconstrained} to \ref{appendix:proofs} along with any other omitted proofs.

\begin{algorithm}[tb]
	\caption{{SEAR (Spatial Expanding Approval Rule)} for Unconstrained Clustering}
	\label{alg:PRF-unconstrained}
	\begin{algorithmic}
		\small
		\STATE {\bfseries Input:}  metric space $\X$ with a distance measure $d:\X \times \X \rightarrow \mathbb{R}^+\cup \{0\}$, a finite multiset $\N\subseteq \X$ of $n$ data points (agents), and positive integer $k$.
		\STATE {\bfseries Output:}  A multiset of $k$ centers
		\STATE $w_i\longleftarrow 1$ for each $i\in \N$			
		\STATE Consider the set $D=\{d(i,i')\mid i,i'\in N\}$. Order the entries in $D$ as $d_1\leq d_2 \leq \cdots d_{|D|}$.
		% \STATE $\M=\{c_i^1,\ldots, c_i^k \mid i\in \N\}$
		\STATE $\M\longleftarrow \N$
		% \barton{Should this be $\M=\{c_i^1,\ldots, c_i^k \mid i\in \N\}$?}
		%   \STATE The location of each $c_i^j$ is the same as agent $i$.
		\STATE $j\longleftarrow 1$
		\STATE $W\longleftarrow \emptyset$
		\WHILE{$|W|<k$}
		%\STATE Draw circles of radii $d_i$ for each $i\in \N$
		\STATE $C^*=\{c\in \M\mid \sum_{i\in \N\mid d(i,c)\leq d_j} w_i\geq n/k\}$
		\IF{{$C^*=\emptyset$}}
		\STATE $j\longleftarrow j+1$
		\ELSE
		\STATE  { Select some candidate $c^*$ from $C^*$ such that $c^*=\arg\max_{c'\in C^*} \sum_{i\in \N\mid d(i,c')\leq d_j} w_i$}:\\
		$W\longleftarrow W\cup \{c^*\}$; $\M\longleftarrow \M\setminus \{c^*\}$
		\STATE {$N'\longleftarrow \{i\in \N\, :\, d(i,c^*)\leq d_j\}$}
		\STATE \label{step-reweighting} % Modify the weights of agents who supported $c$ in the $j$-approval election. Modify the weights of agents in $N'$ so the total weight of agents in $N'$ decreases by $q$.
		Modify the weights of agents in $N'$ {so the total weight of agents in $N'$, i.e., $\sum_{i\in N'} w_i$,} decreases by exactly $n/k$. 
		\ENDIF
		%{	\{Comment: Details of this reweighting process will be provided later.\}					}
		\ENDWHILE
		\STATE Return $W$
	\end{algorithmic}
\end{algorithm}

\begin{lemma}\label{lemma: alg 1 sat PRF}
	Algorithm~\ref{alg:PRF-unconstrained} returns an outcome that satisfies PRF.\footnote{\jvnew{We note that Algorithm~\ref{alg:PRF-unconstrained} does not rely on the triangle inequality, and hence, can be applied in general semimetric spaces. Since the proof argument of \Cref{lemma: alg 1 sat PRF} does not use the triangle inequality, the PRF guarantee is maintained. The same is true of the PRF guarantee of \Cref{alg:PRF-discrete} in \Cref{Section: discrete}. The same cannot be said of the approximate PF guarantees of the algorithms (see \Cref{lem:3pf}), the proof arguments of which do depend on the triangle inequality.}}
\end{lemma}

\begin{proof}
	Suppose for contradiction that the algorithm finds a set of $k$ centers $W$ that violates PRF. 
	In that case, there exist a set of agents $S\subseteq \N$ of size at least $|S| \geq \ell n/k$ such that the maximum distance between pairs of agents in $S$ is $y$ but the number of locations that are within $y$ of some agent in $S$ is at most $\ell-1$. 
	Consider the snapshot of the algorithm when neighborhood distance $y=\max_{i,j\in S}d(i,j)$ is considered. 
	At this point, 
	$|c\in X\midd \exists i\in S \text{ s.t } d(i,c)\leq y|\leq \ell-1$. 
	Since agents in $\N$ have only used their weight towards selecting centers within $y$ up to this point \jvnew{and each selected center reduced the collective weight of $S$ by at most $n/k$}, it follows that $\sum_{i\in S}w_i\geq \ell \frac{n}{k} - (\ell-1)\frac{n}{k}=\frac{n}{k}$. 
	Therefore, the agents in $S$ still have total weight at least $n/k$ to select one more location that is a distance of at most $y$ from some agent in $S$. Hence, at this stage, when the neighborhood distance is $y$, the algorithm would have selected at least one more center within distance $y$ than in the outcome $W$.
\end{proof}

	Lemma~\ref{lemma: alg 1 sat PRF} implies guaranteed existence of an outcome satisfying our axiom, which was one of our goals in devising PRF. 
	In conjunction with Proposition~\ref{prop:approx_pf_unconstrained}, the lemma also gives us that the outcome returned by Algorithm~\ref{alg:PRF-unconstrained} is $\frac{1}{2}(3+\sqrt{17})$-PF. 
	This is noteworthy as it establishes Algorithm~\ref{alg:PRF-unconstrained} as the first known polynomial-time approximation algorithm for PF on general metric spaces in the unconstrained setting.
	As we will show next, Algorithm~\ref{alg:PRF-unconstrained} actually provides the stronger guarantee of $3$-PF. 
	
	\begin{lemma} \label{lem:3pf}
		Algorithm~\ref{alg:PRF-unconstrained} finds an outcome that satisfies $3$-approximate PF.
		\jvnew{There is an instance for which Algorithm~\ref{alg:PRF-unconstrained} does not satisfy $(3-\epsilon)$-PF for any $\epsilon>0.$}
	\end{lemma}
	\begin{proof}
		For an instance of unconstrained clustering, let $X$ be the outcome returned by Algorithm~\ref{alg:PRF-unconstrained}.
		Fix $S\subseteq N, c\in \M$ with $|S| \geq \lceil n/k \rceil$.
		We will show that there exists an agent $i\in S$ and center $x\in X$ such that $d(i,x)\leq 3d(i,c)$.
		We assume that $S\cap X=\emptyset$, since otherwise $d(i,x)=0$ for some agent and center, and we are done.
		
		At the start of the algorithm, the collective weight of agents in $S$ is $|S|$. 
		\jvnew{When the algorithm terminates it has selected a set of $k$ centers and, with each candidate selected, the collective weight of $N$ is reduced by exactly $n/k$. 
		Therefore, when the algorithm terminates, the collective weight of $S$ (and also $N$) is exactly zero.}
		% When the algorithm terminates, the collective weight of agents in $S$ is zero.
		Thus, in some iteration, the collective weight of $S$ decreases for the first time. 
		We denote the centroid selected in this iteration by $x^*$ and denote an arbitrary agent whose weight is decreased in this iteration by $i^*$.
		Let $y$ be the maximum distance between $i^*$ and any other agent in $S$. 
		
		We first point out that $y\geq d(i^*,x^*)$. To see this, suppose for contradiction $y<d(i^*,x^*)$.
		Since $y$ and $d(i^*,x^*)$ are both distances between pairs of agents in $N$, $y=d_j$ and $d(i^*,x^*)=d_{j'}$ for some $j$ and $j'$.
		Furthermore, $j<j'$, since $D$ is in increasing order and by assumption that $y<d(i^*, x^*)$.
		Thus, when the algorithm considers neighborhoods of $d_j=y$, the collective weight of $S$ has not yet decreased, and we have that 
		$$\sum_{i\in \N| d(i,i^*)\leq y} w_i \geq |S| \geq \lceil \frac{n}{k} \rceil \geq \frac{n}{k}.$$
		Therefore, before transitioning to the next element in $D$, the algorithm will either add $i^*$ to $X$ or the weight of some agent in $S$ will decrease. The first possibility contradicts $S\cap X=\emptyset$. For the second possibility, by assumption, it must be that $x^*$ was added to $X$.
		But, since $d(i^*,x^*)>y$, $i^*\notin N'$ and $w_{i^*}$ does not decrease, which gives us a contradiction.
		
		We divide the remainder of the argument into cases. 
		First, consider the case in which $d(i^*,c)\geq \frac{y}{3}.$ It follows that $d(i^*,x^*)\leq y \leq 3 d(i^*,c).$
		If, instead, $d(i^*,c) < \frac{y}{3}$, then consider the agent $j\in S\setminus \{i^*\}$ such that $d(i^*,j)=y$. 
		By the triangle inequality, we have that 
		\begin{align*}
			d(j,c) &\geq y-d(i^*,c) \geq \frac{2y}{3} \\
			&\geq \frac{1}{3}(d(j,i^*) + d(i^*,x^*)) \geq \frac{1}{3}d(j,x^*).
			%	&\geq \frac{1}{3}d(j,x^*).
		\end{align*}
		% $$d(j,c)\geq y-d(i^*,c) \geq \frac{2y}{3} \geq \frac{1}{3}(d(j,i^*) + d(i^*,x^*)) \geq \frac{1}{3}d(j,x^*).$$ 
		\jvnew{
		We now prove the latter half of the statement by giving an example for which our bound is tight.
		Fix $\epsilon>0$ and let $\delta \in (0, \epsilon/3)$.
		Consider an instance on the real line with $k=2$ and six agents at the following locations:
		$$(-3, -1, -1, 0, 1-\delta, 1-\delta) .$$
		Moreover, suppose the fixed tie-breaking order favors the candidate with the greatest absolute value, and further ties are resolved arbitrarily. 
		The algorithm will proceed as follows. The set of candidates will be initialized as the set of agents and each agent will be given a weight of $1/3$. The candidates at $0$ and $1-\delta$ both capture a total weight of $1$ with a radius of $1-\delta$, which is minimal among all candidates. The tie is broken in favor of one of the candidates at $1-\delta$ and the weights of all agents at $0$ and $1-\delta$ are reduced to zero. Next, the candidate at $-3$ is selected, breaking a tie with $-2$.
		Since all agents have a weight of zero, the algorithm returns $X=\{1-\delta, -3\}$.
		Consider the multiset of agents $S=\{-1,-1,0\}$ and the point $c=-1/3$. 
		Note that $|S|\geq \lceil n/k \rceil$ and for each $i\in S$, it holds that $d(i,1-\delta)/d(i,c) \geq 3\cdot (2-\delta) / 2 = 3- 3\delta/2 > 3-\epsilon.$ Thus, $X$ does not satisfy $(3-\epsilon)$-PF. 
		} 
	\end{proof}

We are now ready to prove Theorem~\ref{thm:unconstrained}.

\begin{proof}[Proof of Theorem~\ref{thm:unconstrained}]
	We prove by induction that if the number of candidates selected is less than $k$, then  there is a way to select at least one more center. If the number of candidates selected is less than $k$, there is still aggregate weight  {of} at least $n/k$ on the  {set of all} agents. These agents can get a candidate selected from their neighborhoods if the neighborhood radius is large enough. In particular, for $d_{|D|}=\max_{i,j\in N}d(i,j)$, some unselected candidate from $M$ can be selected. 

	\jvnew{	
	Next, we bound the running time. 
	For each agent $i \in N$, precompute the list $L_i=(j_1,\ldots,j_n)$ of agents in nondecreasing order of their distance from $i$, so that $d(i,j_1)\leq \cdots \leq d(i,j_n)$. 
	This preprocessing takes $O(n^2\log n)$ time.}

	\jvnew{
	In each execution of the outer loop of \Cref{alg:PRF-unconstrained}, the weights are fixed until a new center is selected. Hence, for a fixed candidate center $i$, the smallest radius $r$ such that $\sum_{j\in N:\ d(i,j)\leq r} w_j \geq n/k$ can be found in $O(n)$ time by scanning $L_i$ from left to right and maintaining the corresponding cumulative weight. 
	Computing this radius for every $i\in N$ therefore takes $O(n^2)$ time, after which the algorithm selects an agent attaining the minimum such radius. The subsequent weight update takes $O(n)$ time.
	Since the outer loop is executed exactly $k$ times, the total time spent after
	preprocessing is $O(kn^2)$. Therefore \Cref{alg:PRF-unconstrained} can be implemented in $O(n^2\log n+kn^2)$ time.}
	% Note that for a given radius $d_j$, at most $k$ candidates can be selected. If no more candidates can be selected for a given $d_j$, the next radius $d_{j+1}$ is considered. There are $O(n^2)$ different distances that are to be considered.  
	% These distances are sorted in $O(n^2 \log(n^2))$ time. 
	% For each distance that we consider, we check for each of the $kn$ locations whether they get sufficient support of $n/k$. 
	% It takes time $kn^2$ to check if some location has support $n/k$ for the current neighborhood distance. 
	% If no location has enough support, we move to the next distance. If some location has enough support, we need to select one less location. Therefore, we need to check whether some location has enough support at most $\max(k,n^2k)$ times.	
	% Therefore the running time is $O(n^2 \log(n^2)) +O((kn^2)(n^2k))=O(n^4k^2)$.
	
	The fact that the algorithm gives PRF and 3-PF follows immediately from Lemma~\ref{lemma: alg 1 sat PRF} and Lemma~\ref{lem:3pf}.
\end{proof}

\section{Fairness for the Discrete Setting}\label{Section: discrete}

We presented PRF as a desirable concept for the unconstrained setting.  It is not  {immediate} how to adapt the concept to the discrete setting. {Applying the unconstrained PRF definition to the discrete setting  leads  to the issues of non-existence. On the other hand, some natural attempts to account for the coarseness of the candidate space, $\M$,  can lead to a concept that is too weak. For example, one alternative is to require that  an outcome $X$ be such that if, for any set of agents $S\subseteq \N$ of size at least $|S|\ge \ell n/k$ such that the maximum distance between pairs of agents in $S$ is $y$, the following holds:  $|c\in X\midd \exists i\in S \text{ s.t } d(i,c)\leq y|\geq  \min\{\ell, |\cup_{j\in S} B_y(j)\cap \M|\}$, where $B_r(i)$ denote a ball of radius $r$ around agent $i$. Alternatively, we could replace the right-hand side of the final inequality with  $ \min\{\ell, |\cap_{i\in S} B_y(i)\cap \M|\}$. Both of these versions are too weak.  {To see this, s}uppose $k=1$, all agents are located at $0$, and $\M=(\ldots, -2, -1, 1, 2, 3, \ldots)$, then neither version places any restriction on the facility location and do not even imply UP.}

{To resolve these issues, we  need to take into account that the nearest candidate locations may be very far from a subset of agents.} We adapt our PRF concept for the discrete setting via a careful adaptation of the PRF concept for the unconstrained setting.

\begin{definition}[Proportionally Representative Fairness (PRF) for Discrete Clustering] \label{def:prf_discrete}
	An outcome X with $|X| = k$ satisfies PRF if the following holds.
	For any set of agents $S\subseteq \N$ of size at least $|S|\geq \ell n/k$, if there are $\ell'\leq \ell$ candidates from $\M$ that are at distance at most $y$ from all agents in $S$, the following holds:  $|c\in X\midd \exists i\in S \text{ s.t } d(i,c)\leq y|\geq \ell'$.
\end{definition}

{
	We begin by highlighting the fact that, in the discrete setting, PRF implies $(1+\sqrt{2})$-approximate PF, the closest PF approximation known to exist in the discrete setting \citep{CFLM19a}. Our proof follows similarly to Theorem 1 from \citet{CFLM19a}.}
\begin{proposition}\label{prop:PRF-PF}
	{{For metric spaces,} if an outcome $X\subseteq \M$ with $|X|=k$ satisfies PRF for Discrete Clustering, then $X$ is $(1+\sqrt{2})$-approximate PF.}
\end{proposition}
\begin{proof}%[Proof of Proposition~\ref{prop:PRF-PF}]
	Let $X$ be an outcome which satisfies PRF for Discrete Clustering.
	Fix $S\subseteq N, c\in \M$ with $|S| \geq \ceil{n/k}$.
	%First note that if $d(i,c) = 0$ for any $i\in S$, then $X$ satisfies PF.
	%Thus, we assume $d(i,c)>0$ for all $i\in S$.
	Let $y = \max_{i\in S} d(i,c)$, the maximum distance any agent in $S$ is from $c$, and denote this agent $i^*$. {We assume that $y>0$; otherwise, if $y=0$,  then every agent in $S$ is located at $c$, $c\in X$ (by PRF), and $d(i, c)=d(i, X)$ for all $i\in S$.}
	
	%	 because $X$ satisfies PRF, it must be that $d(i,c)=0$ 
	
	Since \jvnew{$|S|\geq \ceil{n/k} \ge n/k$} and every agent in $S$ is a distance of at most $y$ from $c$, it follows from $X$ satisfying PRF that there exist $x\in X, i\in S$ such that $d(i,x)\leq y$.
	By the triangle inequality, $d(c,x) \leq d(i,c) + d(i,x)$ and thus $d(i^*, x) \leq y + d(i,c) + d(i,x)$. From these observations, the minimum PF approximation ratio obtained by $i$ or $i^*$ becomes
	\begin{align*}
		& \min \left ( \frac{d(i,x)}{d(i,c)}, \frac{d(i^*,x)}{d(i^*,c)} \right) \\
		& \leq \min \left ( \frac{y}{d(i,c)}, \frac{y+ d(i,c) + d(i,x)}{y} \right) \\
		& \leq \min \left ( \frac{y}{d(i,c)}, 2 + \frac{d(i,c)}{y} \right) \\
		& \leq \max_{z\geq 0}( \min (z, 2 + 1/z)) = 1 + \sqrt{2}.
	\end{align*}
	
	Note that the final inequality above holds since replacing $\frac{y}{d(i,c)}$ by the non-negative number $z$, which maximizes the expression, can only weakly increase the expression. It can be easily verified that this expression is maximized when $z=1+\sqrt{2}$, hence the final transition. Thus, for any $c\in \M$, there exists $i\in S$ such that $d(i,X) \leq (1+\sqrt{2}) d(i,c)$.
\end{proof}

For the discrete setting, we propose a new algorithm (Algorithm~\ref{alg:PRF-discrete}). Algorithm~\ref{alg:PRF-discrete} is similar to Algorithm~\ref{alg:PRF-unconstrained}{, which was} designed for the unconstrained setting. In fact, Algorithm~\ref{alg:PRF-unconstrained} can be viewed as first setting $\M$ to the multiset of candidate locations corresponding to agents in $\N$ and then running Algorithm~\ref{alg:PRF-discrete}. 
{Similar to  Algorithm~\ref{alg:PRF-unconstrained}, Algorithm~\ref{alg:PRF-discrete} terminates in polynomial time and returns $k$ centers and its output satisfies PRF.

	\begin{algorithm}[tb]
		\caption{{SEAR (Spatial Expanding Approval Rule)} for Discrete Clustering}
		\label{alg:PRF-discrete}
		\begin{algorithmic}
			\small
			\STATE {\bfseries Input:}  metric space $\X$ with a distance measure $d:\X \times \X \rightarrow \mathbb{R}^+\cup \{0\}$, a finite multiset $\N\subseteq \X$ of $n$ data points (agents), a finite set of candidate locations $\M$, and positive integer $k$.
			\STATE {\bfseries Output:}  A multiset of $k$ centers
			\STATE $w_i\longleftarrow 1$ for each $i\in \N$			
			\STATE Consider the set $D=\{d(i,c)\mid i\in \N,c\in \M\}$. Order the entries in $D$ as $d_1\leq d_2 \leq \cdots \leq d_{|D|}$.
			\STATE $j\longleftarrow 1$
			\STATE $W\longleftarrow \emptyset$
			\WHILE{$|W|<k$}
			\STATE $C^*=\{c\in \M\mid \sum_{i\in \N\mid d(i,c)\leq d_j} w_i\geq n/k\}$
			\IF{{$C^*=\emptyset$}}
			\STATE $j\longleftarrow j+1$
			\ELSE
			\STATE  { Select some candidate $c^*$ from $C^*$ such that $c^*=\arg\max_{c'\in C^*} \sum_{i\in \N\mid d(i,c')\leq d_j} w_i$ }:\\
			$W\longleftarrow W\cup \{c^*\}$;
			$\M\longleftarrow \M\setminus \{c^*\}$
			\STATE {$N'\longleftarrow \{i\in \N\, :\, d(i,c^*)\leq d_j\}$}
			\STATE \label{step-reweighting} 	 Modify the weights of agents in $N'$ {so the total weight of agents in $N'$, i.e., $\sum_{i\in N'} w_i$,} decreases by exactly $n/k$. 
			\ENDIF
			
			\ENDWHILE
			\STATE Return $W$
			
		\end{algorithmic}
	\end{algorithm}
	
	\begin{theorem}\label{thm:discrete}
		Algorithm~\ref{alg:PRF-discrete} terminates in polynomial time \jvnew{$(|M|n\log n + k|M|n)$} and returns a set of $k$ centers which satisfy PRF and $(1+\sqrt{2})$-approximate PF. 
	\end{theorem}
	\begin{proof}[Proof of Theorem~\ref{thm:discrete}]
		\par\underline{Output size.}
		We prove by induction that if the number of candidates selected is less than $k$,
		then there is a way to select at least one more center. If the number of candidates selected is less than $k$,  there is still aggregate weight  {of} at least $n/k$ on the  {set of all} agents. These agents can get a candidate selected from their neighborhood if the neighborhood distance is large enough. In particular, for $d_{|D|}=\max_{i\in N, c\in \M}d(i,c)$, some unselected candidate from $M$ can be selected.
		%
%		\par\underline{Running time.}
%		Note that for a given radius $d_j$, at most $k$ candidates can be selected. If no more candidates can be selected for a given $d_j$, the next distance $d_{j+1}$ is considered.
%		Thus, here are $(n\cdot |\M|)$ different distances to be considered.  These distances are sorted in $O(n\cdot |\M| \log(n\cdot |\M|))$ time. For each distance that we consider, we check for each of the $|\M|$ locations whether they get sufficient support of $n/k$. It takes time $|\M|\cdot n$ to check if some location has support $n/k$ for the current neighborhood distance. If no location has enough support, we move to the next distance. If some location has enough support, we need to select one less location. Therefore, we need to check whether some location has enough support at most $\max(k,n\cdot |\M|)$ times.	
%		Therefore the running time is $O(n\cdot |\M| \log(n\cdot |\M|)) +O((|\M|\cdot n) (k+n\cdot |\M|))=O({|\M|}^2n)$.
%		%

	\par\underline{Running time.}
	\jvnew{	
	% Let $m=|\mathcal M|$.
	For each candidate center $c \in \mathcal M$, precompute the list $L_c=(j_1,\ldots,j_n)$ of agents in nondecreasing order of their distance from $c$, so that $d(c,j_1)\leq \cdots \leq d(c,j_n)$. 
	This preprocessing takes $O(|M|n\log n)$ time.}

	\jvnew{
	In each execution of the outer loop of \Cref{alg:PRF-discrete}, the weights are fixed until a new center is selected. 
	Hence, for a fixed candidate center $c$, the smallest radius $r$ such that $\sum_{j\in N:\ d(c,j)\leq r} w_j \geq n/k$ can be found in $O(n)$ time by scanning $L_c$ and maintaining the corresponding cumulative weight. 
	Computing this radius for every $c\in \mathcal M$ therefore takes $O(|M|n)$ time, after which the algorithm selects a candidate center attaining the minimum such radius. The subsequent weight update takes $O(n)$ time.
	Since the outer loop is executed exactly $k$ times, the total time spent after preprocessing is $O(k|M|n)$. Therefore \Cref{alg:PRF-discrete} can be implemented in $O(|M|n\log n + k|M|n)$ time.}
% Note that for a given radius $d_j$, at most $k$ candidates can be selected. If no more candidates can be selected for a given $d_j$, the next distance $d_{j+1}$ is considered.
% Thus, here are $(n\cdot |\M|)$ different distances to be considered.  These distances are sorted in $O(n\cdot |\M| \log(n\cdot |\M|))$ time. For each distance that we consider, we check for each of the $|\M|$ locations whether they get sufficient support of $n/k$. It takes time $|\M|\cdot n$ to check if some location has support $n/k$ for the current neighborhood distance. If no location has enough support, we move to the next distance. If some location has enough support, we need to select one less location. Therefore, we need to check whether some location has enough support at most $\max(k,n\cdot |\M|)$ times.	
% Therefore the running time is $O(n\cdot |\M| \log(n\cdot |\M|)) +O((|\M|\cdot n) (k+n\cdot |\M|))=O({|\M|}^2n)$.
%

		\par\underline{PRF.}
		Suppose, for the sake of a contradiction, that the algorithm finds a set, $X$, of $k$ centers that violates PRF. 
		In that case, there exists a set of agents $S\subseteq \N$ of size at least $|S| \geq \ell n/k$ such that there are at least $\ell'\leq \ell$ locations in $\M$ within distance $y$ of every agent in $S$, but the number of locations in $X$ that are within $y$ of at least one agent in $S$ is at most $\ell-1$. 
		In that case, consider the snapshot of the algorithm when neighborhood distance $y$ is considered. 
		At this point, the number of locations selected by the algorithm that are within distance $y$ of at least one agent in $S$ is at most $\ell-1$. 
		Since, to this point, agents in $\N$ have only used their weight towards selecting locations within distance $y$, it follows that $\sum_{i\in S}w_i\geq \ell \frac{n}{k} - (\ell-1)\frac{n}{k}=\frac{n}{k}$. 
		It follows that the agents in $S$ still have total weight at least $n/k$ to select one more location that is at distance at most $y$ from some agent in $S$. 
		Hence, at this stage, when the neighborhood distance is $y$, the algorithm would have selected at least one more center within distance $y$ of some agent in $S$. Thus, $X$ is not the correct output of the algorithm, a contradiction.
		\par\underline{PF approximation.}
		Since, as we just showed, the outcome returned by Algorithm~\ref{alg:PRF-discrete} satisfies PRF, it follows from Proposition~\ref{prop:PRF-PF} that the outcome returned by Algorithm~\ref{alg:PRF-discrete} satisfies $(1+\sqrt{2})$-PF.
	\end{proof}
	
	Notably, by satisfying PRF, Algorithm~\ref{alg:PRF-discrete} matches the best known PF approximation factor in the discrete setting. 
	As the following remark highlights, however, the existing algorithms do not capture PRF.
	\begin{remark}	\label{rmk:prf_gc}
		{\citet{CFLM19a} present Greedy Capture. An equivalent algorithm, called ALG$_g$, is also presented by \citet{LLS+21a}.  \jvnew{However, these algorithms do not satisfy unanimous proportionality, and thus also do not satisfy PRF.}} {In fact, for given $k$, both algorithms may fail to output $k$ candidate locations.\footnote{To see this, let $k=3$ and let $N=(0,0,1)$. \citeauthor{CFLM19a}'s Greedy Capture and \citeauthor{LLS+21a}'s ALG$_g$ will  select candidate locations 0 and 1, but will not output a third  location. \jvnew{Since unanimous proportionality requires the selection of two centers at 0 and one center at 1, the algorithms violate the stronger axiom of PRF in both the unconstrained and discrete setting. In this case, re-running Greedy Capture with the remaining centers will give PRF, but this does not hold in general.}}}
		% Of course, this issue could be rectified by arbitrarily choosing a third candidate location---but then there is no guarantee that the set of locations would satisfy PRF. } }
	\end{remark}

	% {In \ref{appendix:experiment}, we experimentally compare Algorithm~\ref{alg:PRF-discrete} to ALG$_g$ and a variant of k-means. 
	% 	We show that it performs especially well when the mean squared distance of the agents is  {calculated based on their distance from not only the closest center but also the next $j$-closest centers (for large $j$).
	% 		This provides another perspective indicating that, in situations where a ``fair'' clustering should account for an agent's distance from more than just her closest center, our algorithm is better suited than the existing alternatives.
			In \ref{app: strengthen PRF}, we consider two natural and stronger notions of PRF in the discrete setting (that also apply {to} the continuous setting). 
			We show that the versions that we consider do not guarantee the existence of a fair outcome.}

	\section{Strategyproofness and Fairness}\label{appendix: SP and fairness} 
		
		Most of the work on clustering assumes that the data points are truthfully reported. If data points correspond to individual agents who prefer to have a nearby center, then the issue of incentives also arises (see, e.g., \citet{Proc08b}). {It is easy to see that Algorithm~\ref{alg:PRF-unconstrained} for the unconstrained setting coincides with the mid-point mechanism if there are two agents on a line. Since the mid-point (egalitarian mechanism) is known to be manipulable~(see, e.g., \citet{PrTe13a}), Algorithm~\ref{alg:PRF-unconstrained} is not strategyproof even for $k=1$ (under which an agent cares about the sole facility to be as close as possible). A discretized version of the same example shows that Algorithm~\ref{alg:PRF-discrete} is also not strategyproof. }
		
		Instead of understanding the strategic aspects of a particular algorithm, we next examine the tradeoff between fairness and strategyproofness. 
		We first focus on the case of $k=1$ for which the preference relation of an agent is clear: an agent wants the location to be as close as possible. 
		
		\begin{proposition}	\label{prop:PF_SP}
			Even for $k=1$ and the unconstrained setting, there is no anonymous, PF, and strategyproof algorithm. 
		\end{proposition}
		\begin{proof}
			For $k=1$ and Euclidean  space, an outcome is Pareto optimal if and only is it is in the convex hull of the agent locations~\citep{PSS92a}.  Next, we show that weak Pareto optimality is equivalent to Pareto optimality in this context.
			We show that if a location $x$ is not Pareto optimal, then it is not weakly Pareto optimal. 
			Since $x$ is not Pareto optimal, it is strictly outside the convex hull $H$ of the agent locations. Let $y$ be the point in $H$ closest to $x$. Note that $y$ will either be a corner of $H$ or the perpendicular projection of $x$ onto an edge of $H$. Consider the line $L$ going through $y$ orthogonal to the line through $x$ and $y$. Assume without loss of generality that $L$ is vertical with $x$ to the right of $L$. Then all points of $N$ lie on $L$ or to the left of $L$.
			Let $i$ be an arbitrary point in $N$.
			If $i$ is the same point as $y$, then $d(i,y)<d(i,x)$. Now suppose $i$ is not at the same location as $y$. Consider the triangle $(i,y,x)$. Since $x$-$y$ is orthogonal to $L$, it follows that the angle at point $y$ for triangle  $(i,y,x)$ is at least $90^{\circ}$. The maximum angle of the other two corners of the triangle $(i,y,x)$ is strictly less than $90^{\circ}$ as the sum of the angles of a triangle is $180^{\circ}$. There is a well-known triangle inequality law that ``\emph{if two angles of a triangle are unequal, then the measures of the sides opposite these angles are also unequal, and the longer side is opposite the greater angle.}'' Hence, it follows that $d(i,y)<d(i,x)$. We have shown that $d(i,y)<d(i,y)$ for all $i\in N$. Hence $x$ is not weakly Pareto optimal.
			
			\citet{PSS92a} proved that there is no anonymous, strategyproof, and Pareto optimal mechanism for the Euclidean space with 3 dimensions. From our argument above, it follows that there is no anonymous, strategyproof, and weakly Pareto optimal mechanism for the Euclidean space with 3 dimensions.
			Since PF is equivalent to weak Pareto optimality for $k=1$, it follows that for $k=1$, there is no anonymous, PF, and strategyproof algorithm. 
		\end{proof}
		
		\begin{corollary}
			Even for $k=1$ and the unconstrained setting, there is no anonymous, core fair, and strategyproof algorithm. 
		\end{corollary}
		
%		Next, we show that even for $k=2$ and the discrete setting, there exists no PRF strategyproof algorithm even in one dimension.
%		
%		\begin{proposition}
%			Even for $k=2$ and the discrete setting, there exists no PRF and strategyproof algorithm even in one dimension.
%		\end{proposition}
%		\begin{proof}
%			Let $\boldsymbol{x}=(-1, 1, 1.2)$ and $\M=(0, 0.9, 1.3)$. Let $k=2$. PRF (for discrete setting) requires that $X=(0, 0.9)$. This follows because the group of agents $S=N$ can demand 2 centers and there exists 2 centers at most $y=2.2$ for all agents locations---namely, $0$ and $0.9$. The other groups of agents do not add any further demands (i.e., no more demands that are not already satisfied by considering $S=N$). With this outcome, agent $3$ has access to 2 facilities with distances $0.3$ and $1.2$, respectively. 
%			
%			Now consider a deviation by agent 3 to $x_3'=1.4$.  Now the group of agents $S=\{2,3\}$ can demand that the candidate center at $1.3$ is chosen---this is because this is the only center that is within $y=0.4$ of both agents 2 and 3.  There are 2 possible outcomes $X'=(0.9, 1.3)$ and $X''=(0, 1.3)$. In the first case, agent $3$ (with true location $1.2$) has cost $0.1$ and $0.3$. In the second case, agent $3$ (with true location $1.2$) has cost $0.1$ and $1.2$. In either case, the new outcome component-wise dominates the original outcome (under sincere voting). Hence, PRF is incompatible with SP in the discrete setting.
%		\end{proof}
%		
	
	\jvnew{We will now prove an incompatibility between PRF and strategyproofness. 
	Since it is not clear whether this incompatibility holds in the special case with $k=1$,\footnote{When $k=1$, PRF is weaker than weak Pareto optimality so \Cref{prop:PF_SP} is not enough to prove an impossibility, even with anonymity.}
	we will introduce a dominance relation in order to rigorously define a weak form of strategyproofness in the setting with arbitrary $k$.
	First, denote an instance of centroid selection with metric space $(\mathcal{X},d)$, candidate set $M\subseteq \mathcal X$, agent set $N\subseteq \mathcal X$, and selection size $k$ as $(\mathcal{X},d,M,N,k)$.
	For an instance of centroid selection $(\mathcal{X},d,M,N,k)$, and an arbitrary agent $i\in N$ and centroid selection $X$, we denote by $\vec{d}(i,X)$ the increasing order statistics of agent $i$'s distances to the centers in $X$, i.e., we set $\vec{d}(i,X) = (d(i,x_1), d(i,x_2),\ldots, d(i,x_k))$ where $X = (x_1,x_2,\ldots,x_k)$ and the indices are chosen so that $d(i,x_1)\leq d(i,x_2)\leq \ldots \leq d(i,x_k)$. 
	Given order statistic $\vec{d}(i,X)$, we denote the $j$-th component by $\vec{d}_j(i,X)$ for $j\in \{1, \ldots, k\}$. Given two  order statistics $\vec{d}(i,X)$ and $\vec{d}(i,Y)$, we use the inequality  $\vec{d}(i,X)\leq \vec{d}(i,Y)$ to denote the component-wise ordering:  $\vec{d}(i,X)\leq \vec{d}(i,Y) \iff \vec{d}_j(i,X)\leq \vec{d}_j(i,Y)$ for all $j\in \{1, \ldots, k\}$.}

	\jvnew{Then, given centroid selections $X$ and $Y$ of size $k$ for a given instance with target size $k$, for any agent $i\in N$, we say that $X$ \emph{order-statistic dominates} $Y$  with respect to agent $i$ (denoted $X\triangleright_i Y$) if $\vec{d}(i,X)\leq \vec{d}(i,Y)$ and $\vec{d}_j(i,X)< \vec{d}_j(i,Y)$ for at least one $j\in \{1, \ldots, k\}$. 
	Note that $X\triangleright_i Y$ provides a very strong argument that $i$ should prefer $X$ to $Y$.
	In particular, it implies that any utility function over centroid selections monotonically non-increasing with respect to the individual distances from the agent to the selected centers would evaluate $X$ higher than $Y$.
	This formulation of an incomplete order over centroid selections is in the spirit of weak strategyproofness as studied in randomized social choice with ordinal preferences (see, e.g., \citet{BrLe25a}). 
	In that work, agents' prefer one outcome to another only if it improves their expected utility for all additive utility functions consistent with the agent's ordinal preferences.}
	% a deviation is regarded as profitable only if it leads to an outcome that is unambiguously better for the agent with respect to a prescribed class of admissible utility functions. 
	 % improve expected utility for all utility functions consistent with the agent's ordinal preferences, 
	 % This is in contrast to the stronger classical notion in that setting that already counts improvement for one such utility function as a successful manipulation. 
	 % Our notion is analogous: agents prefer one selection to another only if the resulting committee strictly improves the agent's sorted distance vector in the product order.

	\jvnew{Using our strong requirement for a preference relation then gives us a very weak definition of strategyproofness, strengthening our forthcoming impossibility. 
	Notice that, when $k=1$, our definition of strategyproofness reduces to the standard notion, as used earlier in this section.}

	\begin{definition}
		\jvnew{A centroid selection algorithm $F$ satisfies \emph{order-statistic strategyproofness} if, for every instance of centroid selection $I=((\mathcal{X},d), N, M, k)$, every agent $i\in N$, and every misreport $i'\in \mathcal{X}$, it holds that $F(I) \not \triangleright_i F(I')$ (i.e., $F(I)$ is not order-statistic dominated by $F(I')$) where $I' = ((\mathcal{X},d), N\setminus \{i\}\cup\{i'\}, M, k)$.}
	\end{definition}

	\jvnew{We will prove that no centroid selection algorithm can satisfy both order-statistic strategyproofness and PRF. 
	To do so, we will use a reduction from approval-based committee voting, the notation of which we will now introduce. 
	An instance of committee voting is described by a candidate set $C$, a committee size $k$, and an approval profiles $A=(A_1, A_2, \ldots, A_n)$, which specifies an approval set $A_i\subseteq C$ for each voter $i$. 
	Our reduction will use a target selection size of $k$ and $n$ agents, which is why we do not rename these parameters. 
	The utility of voter $i$ for a selected committee $W\subseteq C$ of size $k$ is equal to the number of selected candidates they approve, i.e., $u_i(W) = |A_i\cap W|.$
	A voting rule maps from approval profiles to committees of size $k$ and is strategyproof if no agent $i\in N$ can misreport their approval set and obtain a strictly higher utility.
		We now prove an impossibility between PRF and order-statistic strategyproofness by showing that, if a PRF and order-statistic strategyproof algorithm exists, then there exists an approval-based committee rule satisfying strategyproofness and PJR, contradicting a known impossibility \citep{Pete18a}.}
	\begin{theorem}
		\jvnew{Even when $k=2$, there is no order-statistic strategyproof algorithm that satisfies PRF in the discrete setting.}\footnote{\jvnew{The statement holds for $m\ge 4$ and even $n$, in accordance with the assumptions used by \citet[][Theorem 5.8]{Pete18a}}.}
		% Let $F$ be an algorithm for discrete metric committee voting on arbitrary finite metric spaces.
		% If $F$ is order-statistic strategyproof and satisfies PRF, then there is an approval-based committee rule $\widehat F$ that is strategyproof and satisfies PJR.
		% Consequently, no such rule $F$ exists.}
	\end{theorem}
	\begin{proof}
		\jvnew{Fix a candidate set $C$. 
		We begin by defining a reduction from an instance of approval-based committee voting with approval profile $A$ and committee size $k$ to an instance of centroid selection.
		First, define the set
		\[
		\mathcal{X}_C := C \cup \{p_T : T \subseteq C\}.
		\]
		Define a weighted bipartite graph on $\mathcal X_C$ by adding an edge $(p_T,c)$ of length $1$ if
		$c \in T$ and of length $2$ otherwise, and let $d_C$ be the shortest-path metric.
		For every $T \subseteq C$ and $c \in C$, the direct edge is shortest, so
		$d_C(p_T,c)=1$ if $c\in T$ and $d_C(p_T,c)=2$ otherwise.}

		\jvnew{Given an approval profile $A=(A_i)_{i\in N}$, define the metric instance
		\[
		I(A):=(\mathcal{X}_C,d_C,C,\{p_{A_i}:i\in N\},k).
		\]
		That is, our reduction uses the metric space defined by $(\mathcal{X}_C, d_C)$, sets the points in $C$ as the set of candidate centers, and each agent $i$ from the committee voting instance is mapped to a single point $p_{A_i}$ which is at distance $1$ from candidates which $i$ approves and distance $2$ from all other candidates. 
		For every voter $i$ and size-$k$ centroid selections $X,X' \subseteq C$, we have that 
		\begin{equation} \label{eq:pref_equivalence}
			X\triangleright_{p_{A_i}} X' \quad \iff \quad |\{x\in X: d(p_{A_i},x)=1\}|>|\{x\in X': d(p_{A_i},x)=1\}| \quad \iff \quad |X\cap A_i| > |X'\cap A_i|.	
		\end{equation}
		%
		% $d_C(p_{A_i},)=1$ iff $W\cap A_i\neq\varnothing$, and otherwise $d_C(p_{A_i},W)=2$.
		Hence, the induced preferences over center selections coincide exactly with the preferences over committees.}

		\jvnew{
		Suppose, toward a contradiction, that there exists a centroid selection rule, $F$, that satisfies order-statistic strategyproofness and PRF on all instances with $k=2$, $m\ge 4$, and even $n$.
		Define the voting rule $\widehat F$ which first follows the reduction to center selection and then executes $F$ on the reduced instance and returns the committee corresponding to the result, i.e., $\widehat F(A):=F(I(A))$. 
		If $\widehat F$ were manipulable by voter $i$ from $A_i$ to $A'_i$, then in the fixed metric space $(X_C,d_C)$, the agent located at $p_{A_i}$ could profitably misreport as $p_{A'_i}$, contradicting the order-statistic strategyproofness of $F$.   
		This follows immediately from \Cref{eq:pref_equivalence}, and thus $\widehat F$ is strategyproof.}

		\jvnew{Now let $S\subseteq N$ be a group of voters such that, for some integer $\ell\leq k$, it holds that $|S|\ge \ell n/k$ and $|\cap_{i\in S} A_i|\ge \ell$.
		This means there are at least $\ell$ candidates in $C$ at distance at most 1 from the agent set $\{p_{A_i}\}_{i\in S}$, and moreover, this set contains $|S|\ge \ell n/k$ agents. 
		Thus, by PRF, the committee $X=\widehat F(A)$ satisfies
		\begin{align*}
			&|x\in X: \exists p_{A_i}\in S \text{ s.t. } d(p_{A_i},x)\leq 1|\geq \ell \iff \\	
			&|x\in X: \exists i\in S \text{ s.t. } x\in A_i|\geq \ell,
		\end{align*}
		which is exactly the definition of proportional justified representation (PJR) from the literature on approval-based committee voting \citep{SFF+17a}.}

		\jvnew{
		Therefore, $\widehat F$ is strategyproof and PJR for approval-based multiwinner voting instances with $k=2$, $m\geq 4$ and even $n$. 
		However, an impossibility from \citet[][Theorem 5.8]{Pete18a} for approval-based multiwinner voting proves that such a rule cannot exist.
		In more detail, \citeauthor{Pete18a} studies the compatibility of a proportionality notion weaker than PJR with a weak notion of strategyproofness.
		A corollary of their result is the following: for any approval-based multiwinner voting rule satisfying PJR, there exists instances  with $k=2$, $m\geq 4$ and even $n$, where an agent, by reporting a subset of their (true) approval set $A_i'\not\subset A_i$, can obtain a committee outcome $X'$ that has a strict superset of approved candidates compared to that obtained via truthful reporting $X$, i.e., $X\cap A_i\subsetneq X'\cap A_i$. 
		Hence, $\widehat F$ being strategyproof and PJR yields a contradiction, implying that when $k=2, m\ge 4$, and $n$ is even, no PRF and order-statistic strategyproof rule exists in the discrete setting.}
		% Such a violation immediately implies a violation of order-statistic strategyproofness. 
		% Hence, no order-statistic strategyproof and PRF rule exists for centroid selection in the discrete setting with general metric spaces.}
	\end{proof}
	
%	\barton{I wonder if this proof argument also work for the unconstrained setting? E.g., Given committee voting instance, do the same construction but introduce distances $1$ between each agent and every other agent and arbitrarily large distances between any agent or candidate and any point that doesn't correspond to one of the candidates in $C$?}
%
%	\jv{The tricky part here seems to be that the requirement for a group to constitute a blocking coalition in PRF is different in the unconstrained setting. 
%	If we set a distance 1 between every pair of agents, then every $\ell$-large group of agents is also $\ell$-cohesive at distance 1 and so it seems like a PRF rule might be forced to select agents as centers, not just candidates. 
%	But, if we increases distances above 1 to force the selection of candidates, then there are no cohesive groups at distance 1 so satisfying PRF does not enforce PJR in the voting instance.}
		
		% \vspace{-0.7em}
		\section{Discussion}
		
		{We revisited a classical AI problem with a novel solution concept that builds on ideas from social choice and fair division. 
			We proposed a natural fairness concept called PRF for clustering. PRF is guaranteed to exist and has   desirable features: it implies unanimous proportionality, does not require the specification of protected groups, and is robust to outliers and multiplicative scaling.
			Even if our only focus is on the PF axiom, we make significant progress on efficient algorithms for approximate PF for general metric spaces. 
			{%There are several interesting avenues for future research. 
				We did not focus on strategic aspects. 
				Unfortunately, PRF and PF are incompatible in general with the incentive of an agent to report their location truthfully (see \Cref{appendix: SP and fairness}). 
				A natural next step is to  identify  stronger versions of PRF that still guarantee existence.}    

			\jvnew{The axiomatic results for the algorithms we introduce do not depend on the manner in which agents are reweighted after selecting centers.
			Naturally, we believe that for different applications --- and for different objectives that one is trying to achieve alongside proportionality concerns --- some reweighting schemes may be preferable to others.
			It will be interesting to consider refinements of our algorithms which make more clever {choices} in this regard toward other natural objectives and/or axioms.} 
			Another direction for future research is to understand how much impact the constraint of PRF imposes on standard optimization objectives such as for $k$-means, $k$-median, or $k$-center.

			\section*{Acknowledgments}
			Haris Aziz is supported by the NSF-CSIRO project on ``Fair Sequential Collective Decision-Making''.
			We thank John Dickerson, Bo Li and reviewers of WINE 2024  for helpful comments. 
			%We also thank the reviewers of WINE 2024 for their  
			%*\footnotetext{*\ We owe thanks to John Dickerson and Bo Li for helpful comments and feedback.}
			
			\newpage
			\bibliographystyle{elsarticle-harv}

			\clearpage
			\appendix

		\section{Strengthening of PRF} \label{app: strengthen PRF}
		
		{We consider two natural and stronger notions of PRF in the discrete setting (that also apply  {to} the continuous setting). We will show that the versions that we consider do not guarantee the existence of a fair outcome in all settings.}

		The PRF concept can be increasingly strengthened in the following ways by making the requirements on the outcome stronger.
		
		\begin{definition}[Proportionally Representative Fairness (PRF)-II  for Discrete Clustering]
			For any set of agents $S\subseteq \N$ of size at least $|S|\geq \ell n/k$, if there are $\ell'\leq \ell$ candidates from $\M$ that are at distance at most $y$ from all agents in $S$, the following holds: there exists some $i\in S$ such that  $|c\in X\midd d(i,c)\leq y|\geq \ell'$.
		\end{definition}

		\begin{definition}[Proportionally Representative Fairness (PRF)-III for Discrete Clustering]
			For any set of agents $S\subseteq \N$ of size at least $|S|\geq \ell n/k$, if there are $\ell'\leq \ell$ candidates from $\M$ that are at distance at most $y$ from all agents in $S$, the following holds: $|c\in X\midd d(i,c)\leq y \text{ for all } i\in S|\geq \ell'$.
		\end{definition} 
		
		The following example shows that an outcome satisfying PRF-II and hence PRF-III may not exist.

		\begin{example}
			Take $n=4, k=2$ with the following location profile on a unit interval $[0,1]$:
			$N=(0, 0, 1, 1).$
			There are 4 candidate locations at $\M=(0, 0.5, 0.5, 1)$. It is immediate that the 2 groups of agents at $0$ and $1$ must have a facility at $0$ and $1$. But this violates PRF-II. The set of agents $N$ (size 4) have 2 candidate locations with a distance of $0.5$ of all agents; however, no agent in $N$ has 2 facility locations within a distance of 0.5 of them.
		\end{example}

		\section{\jvnew{Failure of Method of Equal Shares to Satisfy PRF}}
		\label{app:mes_violates_prf}
		\jvnew{In this section, we will show that, to satisfy PRF for discrete clustering, it is not sufficient to transform a clustering instance into an instance of multiwinner voting by setting utilities to be the inverse of distances, and running Method of Equal Shares (MES) \citep{PPS21a}.
		We start by introducing some new notation and giving a brief overview of MES.
		We let $u_i(x) = 1/d(i,x)$ denote the utility agent $i\in \N$ derives from the selection of center $x\in\M$ and set $u_i(x) = \infty$ if $d(i,x)=0$.
		MES first initializes a budget of $b_i=k/n$ for each agent $i$. 
		A candidate $x$ is said to be $\rho$-affordable if $\sum_{i\in N}\min(b_i, \rho\cdot u_i(x)) \geq 1$
		The algorithm proceeds by selecting the candidate $x$ which is $\rho$-affordable for minimum value $\rho$.}

		\jvnew{The candidate $x$ is added and each agent pays $\min(b_i, \rho\cdot u_i(x))$ and their budgets are updated.
		That is, subject to each agent paying equal per-utility price up to budget constraints, the algorithm selects the candidate which can be afforded with the lowest per-utility price. The algorithm continues in this fashion iteratively until $k$ centers are selected.
		We will now give an example for which MES fails to provide PRF when executed on a transformed clustering instance, as above.}

		\begin{example}
			\jvnew{Consider the following instance on the line with five agents, three candidate centers, and $k=2$:
			\begin{itemize}
				\item Three agents located at $-1.$
				\item Two agents located at $1$.
				\item Candidate $a = -9/5$, candidate $x=0$, and candidate $b=8/5$. 
			\end{itemize}
			Each agent is initialized with a budget of $k/n=2/5.$ Each agent $i$ located at $-1$ has utilities $u_i(a) = 5/4, u_i(x) = 1, u_i(b) = 5/13$.
			Each agent $j$ located at $1$ has utilities $u_j(a) = 5/14, u_i(x) = 1, u_i(b) = 5/3.$
			The candidate $x$ is $\rho$-affordable with $\rho=1/5$ since $\sum_{i\in N}\min(2/5, u_i(x)/5) = 3(1/5) + 2(1/5) = 1.$
			One can check that neither $a$ nor $b$ are $\rho$-affordable with $\rho=1$.
			Thus, $x$ is selected first and each agent pays $1/5.$
			Next, $b$ is the candidate which will be $\rho$ affordable for minimum $\rho.$ To see why this is true, note that all of the remaining budget needs to be used to buy the next candidate. Thus, the $\rho$ value for each candidate will be the value required to make the minimum binding for the farthest agent. 
			That is, $a$ is $\rho$-affordable for $\rho\geq 14/25$ and $b$ is $\rho$-affordable for $\rho\geq 13/25.$}

			\jvnew{Now observe that $\{x,b\}$, the center selection made by MES, violates PRF.
			To see this, let $S$ denote the three agents located at $-1$, and note that $|S|\geq n/k $.
			Since $d(i,a)=4/5$ for all $i\in S$, PRF requires that at least one selected center be distance at most $4/5$ away from $x$.
			However, the closest selected center is at distance $1$ from each agent in $S$.}
		\end{example}

\end{document}